 \documentclass[sigconf]{acmart}

\usepackage{booktabs} 
\usepackage[T1]{fontenc}
\usepackage{todonotes}
\usepackage{amsmath}
\usepackage{amssymb}
\usepackage{hyperref}
\setcitestyle{square}

\def\BibTeX{{\rm B\kern-.05em{\sc i\kern-.025em b}\kern-.08emT\kern-.1667em\lower.7ex\hbox{E}\kern-.125emX}}
\newcommand{\norm}[1]{\left\lVert#1\right\rVert}
\copyrightyear{2019} 
\acmYear{2019} 
\setcopyright{acmcopyright}
\acmConference[ETRA '19]{2019 Symposium on Eye Tracking Research and Applications}{June 25--28, 2019}{Denver , CO, USA}
\acmBooktitle{2019 Symposium on Eye Tracking Research and Applications (ETRA '19), June 25--28, 2019, Denver , CO, USA}
\acmPrice{15.00}
\acmDOI{10.1145/3314111.3323073}
\acmISBN{978-1-4503-6709-7/19/06}

\acmSubmissionID{191}

\begin{document}


%
\title{Task Classification Model for Visual Fixation, Exploration, and Search}

%
\author{Ayush Kumar}
\affiliation{%
  \institution{Stony Brook University}
}
\email{aykumar@cs.stonybrook.edu}

\author{Anjul Tyagi}

\affiliation{%
  \institution{Stony Brook University}
}
\email{aktyagi@cs.stonybrook.edu}

\author{Michael Burch}
\affiliation{%
  \institution{Eindhoven University of Technology}
}
\email{m.burch@tue.nl}

\author{Daniel Weiskopf}
\affiliation{%
  \institution{University of Stuttgart}
}
\email{Daniel.Weiskopf@visus.uni-stuttgart.de}

\author{Klaus Mueller}
\affiliation{%
  \institution{Stony Brook University}
}
\email{mueller@cs.stonybrook.edu}

%
\renewcommand{\shortauthors}{Kumar \& Tyagi et. al.}

%
\begin{abstract}
 Yarbus' claim to decode the observer's task from eye movements has received mixed reactions. In this paper, we have supported the hypothesis that it is possible to decode the task. We conducted an exploratory analysis on the dataset by projecting features and data points into a scatter plot to visualize the nuance properties for each task. Following this analysis, we eliminated highly correlated features before training an SVM and Ada Boosting classifier to predict the tasks from this filtered eye movements data. We achieve an accuracy of $95.4\%$ on this task classification problem and hence, support the hypothesis that task classification is possible from a user's eye movement data. 

\end{abstract}

%
%


%

%
\begin{teaserfigure}
  \includegraphics[width=\textwidth]{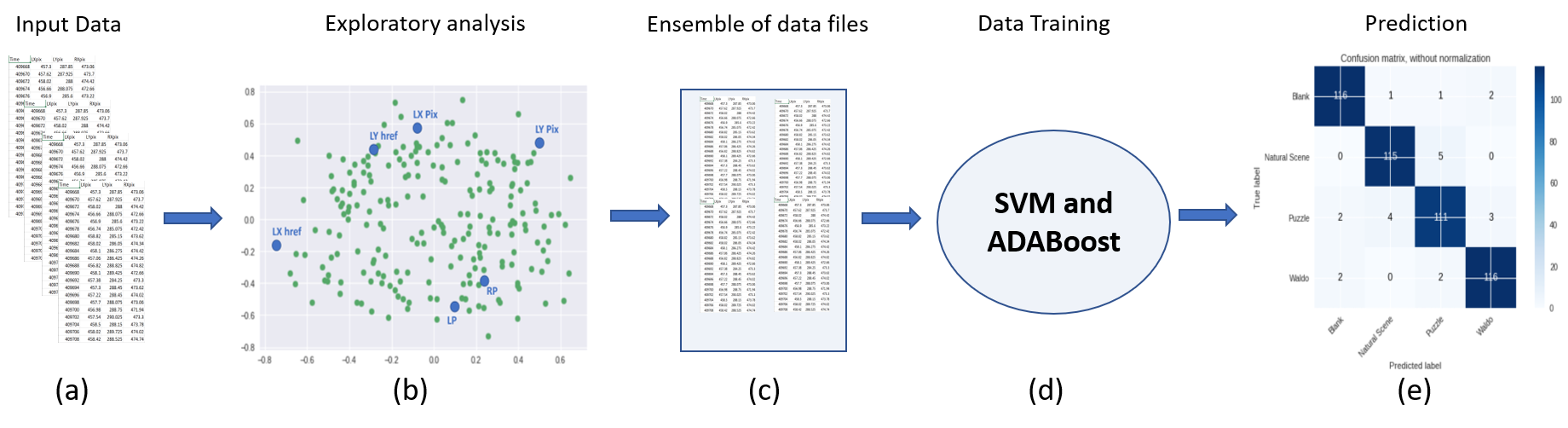}
  \caption{Overview of the classifier trained for task prediction. (a) Combine and shuffle the input files for training in the next stage of exploratory analysis. (b) Feature selection to be done in this stage. (c) Feed the task-specific user file with selected features into the trained classifier. (d) Classifier predictions are analyzed in the form of a confusion matrix shown in (e).}
  \label{fig:teaser}
\end{teaserfigure}

%
\maketitle

\section{Introduction}

Visual attention is one of the most sought area in the field of computer vision and is used in day to day tasks. Extensive studies have been conducted to predict users' behavior or pattern from the visual attention of the observers. Kumar et al.~\cite{kumar2018visual} used multi-metric grouping of eye movements to find out subjects leading to distinct visual groups of similar behavior. Yarbus'~\cite{yarbus2013eye} claim of predicting the observer's task is one such attempt to use visual attention as a cue to look deep into human cognition. Yarbus was the first one to establish the relationship between eye movements and human cognition, which he formed as a basis of his claim. However, Yarbus' claim to decode the observer's task from eye movements has received mixed reaction. Green et al.~\cite{greene2012reconsidering} have reconsidered Yarbus' work and argued against his claim, whereas Borji et al.~\cite{borji2014defending} defended Yarbus and supported their claim with their study. In this paper, we try different classification algorithms to support Yarbus' claim. The results show that task classification is actually possible to predict from eye movement data with an accuracy of 95\% using modern classification techniques in machine learning. To support this claim, we have used a dataset from an extensive study carried out by Otero-Millan et al.~\cite{otero2008saccades}.

The rest of the paper is organized as follows: Section 2 discusses the exploratory analysis part, where we use Data Context Map\cite{cheng2016data} to visualize task specific details in the dataset and for feature selection. Then, Section 3 discusses the proposed classifiers to classify the task on the basis of their eye movements. Section 4 presents the result in the form of a confusion matrix which contains accuracy results. Finally, Section 5 discusses the future work and gives concluding remarks.

\section{Visualization with Data Context Map}

\begin{figure}[h]
    \includegraphics[width=0.45\textwidth]{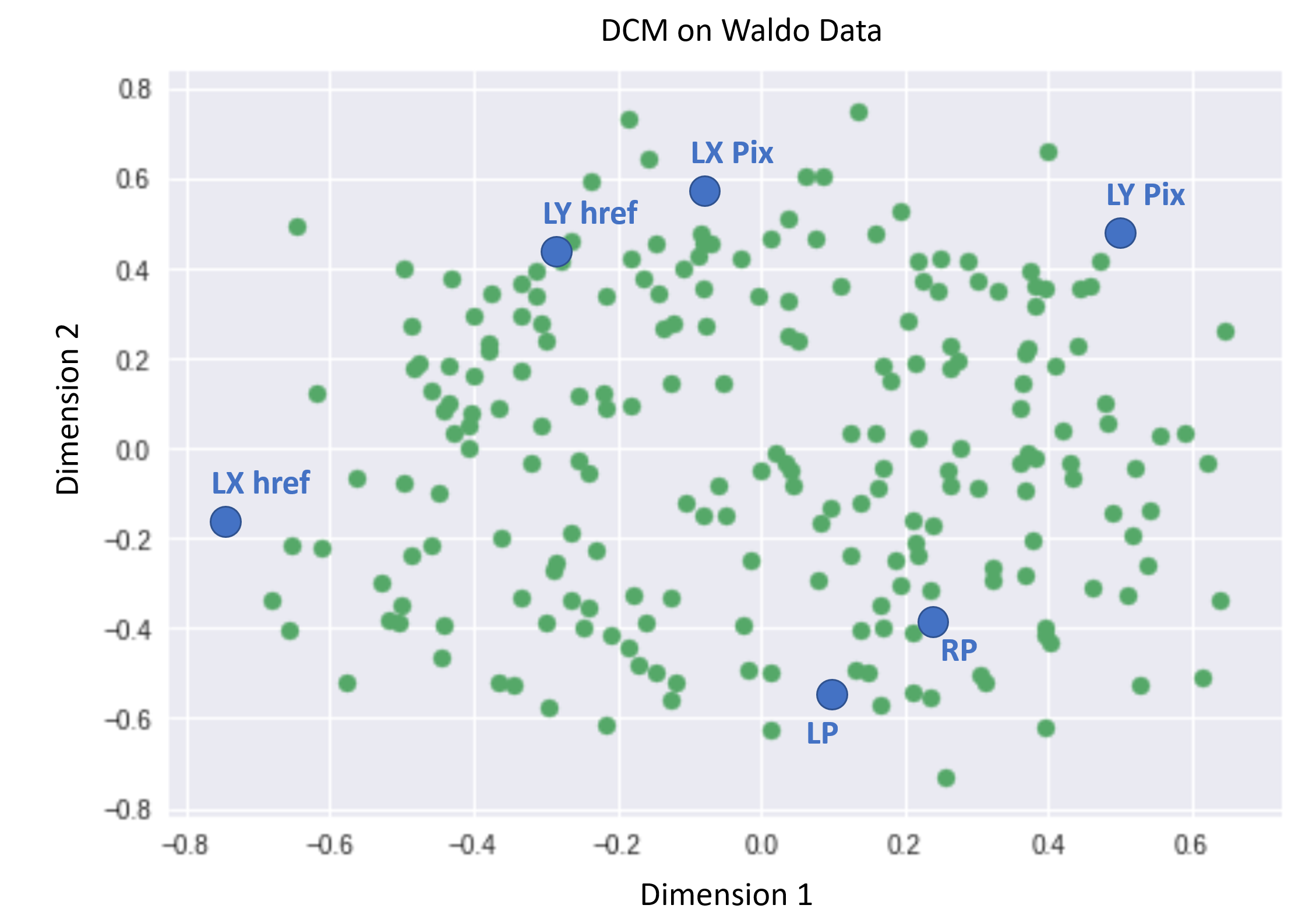}
    \caption{Data Context Map~\cite{cheng2016data} projection of Finding Waldo task on a randomly chosen user trial. Green points represent mapping of eye movement readings on a particular timestamp. Blue points represent corresponding variables in the dataset. }
    \label{fig:dcm}
\end{figure}

The data context map (DCM)~\cite{cheng2016data} is a tool built to visualize high-dimensional data in the form of a 2D scatter plot. DCM is a variation of the multiple correspondence analysis technique described in~\cite{tyagiroad} and is used for numerical data. DCM projection tries to preserve the correlation between variables and the data points at the same time. The plot consists of both, variables and data points represented in a scatter plot and the relative distance between the variables and the data points signifies correlation. An example DCM for variables 
LP, RP, LX href, LY href, LX Pix and LY Pix is shown in Figure \ref{fig:dcm}. The data context map uses multidimensional scaling (MDS)~\cite{kruskal1964multidimensional} to project the data points and the variables into lower dimensions. The distance matrix in MDS corresponds to the correlations in case of inter variable correlation and Euclidean distances in case of inter data correlation. For finding correlation between variables and the datapoints, the variables are treated as a data point with value 1 in the corresponding dimension and zero for all other dimensions. All of these distances are then fused into a single matrix and the matrix is normalized before projecting with MDS. 

\subsection{Analysis}

Data context map can point out subtle relationships between variables and the dataset. We evaluated Data Context Maps for each of the tasks in the datasets to point out interesting correlations and for feature selection. An example representation is shown in Figure \ref{fig:dcm} where we show the context map on \textit{Finding Waldo} task. We can see the that \textit{LP (Left Pupil)} and \textit{RP (Right Pupil)} are close to each other in the projection. This shows that LP and RP have a high correlation. It is also interesting to note that \textit{LX Pix} and \textit{LX href} readings show opposite relations to that of \textit{LP} and \textit{RP} as they show no correlation, since being located far from each other on the context map.

Considering the positioning of data points, it is interesting to note that very few data points lie close to the \textit{LX href} on the context map in Figure \ref{fig:dcm} . This shows that the \textit{LX href} readings had the least correlation with the data points. This can be understood with the concept of randomness in a variable, as more random the distribution of a variable is, the less is it's correlation with respect to a set of data points. This analysis not only allows for visualizing inherent correlations, but also aids in variable filtering. After carefully analyzing the Data Context Maps of all the four tasks, we found out that the least five correlated variables are \textit{LX Pix, LY Pix, LX href, LY href and LP }, which we later used to train the classifiers

\section{Learning Task Classification}

In this task classification, we designed our study to classify four tasks from the fixation dataset. We choose gaze data along with the pupil diameter as analyzed from the the data context map visualizations for training the classifiers. As part of data preprocessing, we created an ensemble of the data by merging the files of all tasks and for all users. Each row of this merged file, which is a reading of eye movements of a user for a specific task at a timestamp was labelled with the corresponding task label. After shuffling all these rows with labels, we generated the training data where each datapoint is the eye movement reading of a user at a timestamp and the label is the label for the task. After the files are merged the dataset is standardized by calculating the z-score for each column in the dataset. As shown in Equation ~\ref{eq:zscore}, zscore is a method to convert a data distribution to standard normal by subtracting the mean $\mu$ and dividing by the standard deviation $\sigma$ for each column distribution in the dataset.

\begin{equation}
  z = \frac{x - \mu}{\sigma}
  \label{eq:zscore}
\end{equation}

We used zscore instead of Min Max normalization since zscore is more robust towards scaling the outliers. Now that the data is normalized, we shuffle the data points and split it to train, test, and validation sets. The testing and validation sets contain about fifteen percent of the total data points each in our experiment. A portion of this normalized data was used to train the two classifiers, \textit{SVM classifier} and \textit{Ada Boosting classifier with decision tree as the base classifier.} By using the readings of \textit{LXPix, LYPix, LX href, LY href, and LP} we could train classification models which can predict the type of task being performed, given the user's readings over time for any of the tasks. 

\subsection{Method: SVM Classifier}

Support Vector Machines~\cite{cortes1995support} is a well studied classifier built to work with binary classification tasks. The main objective is to find the hyper plane which can separate the data points. The data points can be projected to higher dimensions for easier separation using the kernel function. For multiclass classification, one vs. rest classification ~\cite{bottou1994comparison} technique is designed to use multiple binary classifiers to be trained for every label in the dataset~\cite{hsu2002comparison}. Thus, in this case, four different SVM classifiers are trained to each of the four task classifications. The $i^{th}$ SVM is trained with the samples of the $i^{th}$ class as positive and all others as negative. For the given training data $(x_{1}, y_{1}),...,(x_{l}, y_{l})$ with the labels in class from \textit{1} to \textit{k}, the multiclass SVM solves the problem shown in Equation ~\ref{eq:svm}.

\begin{equation}
\begin{aligned}
    \underset{\omega^{i},b^{i},\epsilon^{i}}{\text{min}} \quad &\frac{1}{2} \big(\omega^{i}\big)^{T}\omega^{i} + C\sum_{j=1}^{l} \epsilon_{j}^{i} \big(\omega^{i}\big)^{T}\\
    &\big(\omega^{i}\big)^{T}\phi\big(x_{j}\big) + b^{i} \geq 1 - \epsilon_{j}^{i} &  y_{j} = i\\
    &\big(\omega^{i}\big)^{T}\phi\big(x_{j}\big) + b^{i} \leq -1 + \epsilon_{j}^{i} & y_{j} \neq i\\
    &\epsilon_{j}^{i} \geq 0 & j = 1, ..., l
\end{aligned}
  \label{eq:svm}
\end{equation}

Minimizing the above function separates the positive and negative labels by maximum margin which can be controlled with the cost term C. In our experiment, the value of C = 1,000 gave the best accuracy of around 80\%. We trained SVM on 77,000 data points sampled from the merged dataset with stratified sampling to keep the number of samples from each task class consistent with other tasks. $\phi$ is the kernel function which is used to map the data points from lower to higher dimensions for a better separation of the data points. In our experiment, we use the RBF kernel shown in Equation~\ref{eq:rbf} which is a linear combination of non-linear interpolations of the input to achieve the highest accuracy with SVMs on this task. The trained SVM classifier is used to predict the task from user trials by taking the mode of predicted classes for all samples. 

\begin{equation}
\begin{aligned}
  f(x)=\sum_{i=1}^{n} \alpha_i g(\norm{x-x_{i}}).
 \end{aligned}
  \label{eq:rbf}
\end{equation}

\subsection{Method: Adaptive Boosting}

Adaptive Boosting~\cite{freund1996experiments} is one of the boosting techniques where the main idea is to use an ensemble of classifiers to train on the same dataset. It is based on the concept that training many simple classifiers of accuracy above 50\%, the majority voting of these simple classifiers for each data point is likely to produce better classification results than a complex single classifier. Each classifier is trained to perform better in classifying the training data points which were misclassified by the previous classifiers. Initially, the probability of picking each sample from the dataset is set to $1/N$. After each iteration, when a trained classifier is added to the ensemble, the probability of generating each sample for the next classifier is recalculated. Let $\epsilon_{k}$ be the sum of probabilities of all the misclassified instances for the classifier \textit{$C_{k}$}. Then for the classifier \textit{$C_{K+1}$}, the probability of picking the incorrectly classified samples by \textit{$C_{k}$} is increased by a factor of $\beta_{k} = (1-\epsilon_{k})/\epsilon_{k}$. These probabilities are then normalized so that the sum of probabilities equals 1~\cite{opitz1999popular}. For the base classifier in Ada Boosting, we used Decision Trees with a max depth constraint equal to 6. Trained Ada Boost classifier is used to predict the task from user trials by taking the mode of predicted classes for all samples. Training an ensemble of 100 classifiers with Ada Boosting on 77,000 data points resulted in the accuracy range of 93\% - 95\% in the user task classification. Accuracy comparison between SVM and ADA Boosting is shown in Table ~\ref{tab:accuracy}.

\begin{table}
  \caption{Accuracy of user task classification}
  \label{tab:accuracy}
  \begin{tabular}{ccl}
    \toprule
    Model&Accuracy&Parameters\\
    \midrule
    SVM & 31.2\%& Linear Kernel with C = 1,000\\
    SVM & 80.3\%& RBF Kernel with C = 1,000\\
    \textbf{Ada Boosting} & \textbf{95.4\%} & Number of estimators = 100,\\
    && decision tree max depth = 6\\
  \bottomrule
\end{tabular}
\end{table}

\section{Results}

\begin{figure}[h]
  \centering
  \includegraphics[width=\linewidth]{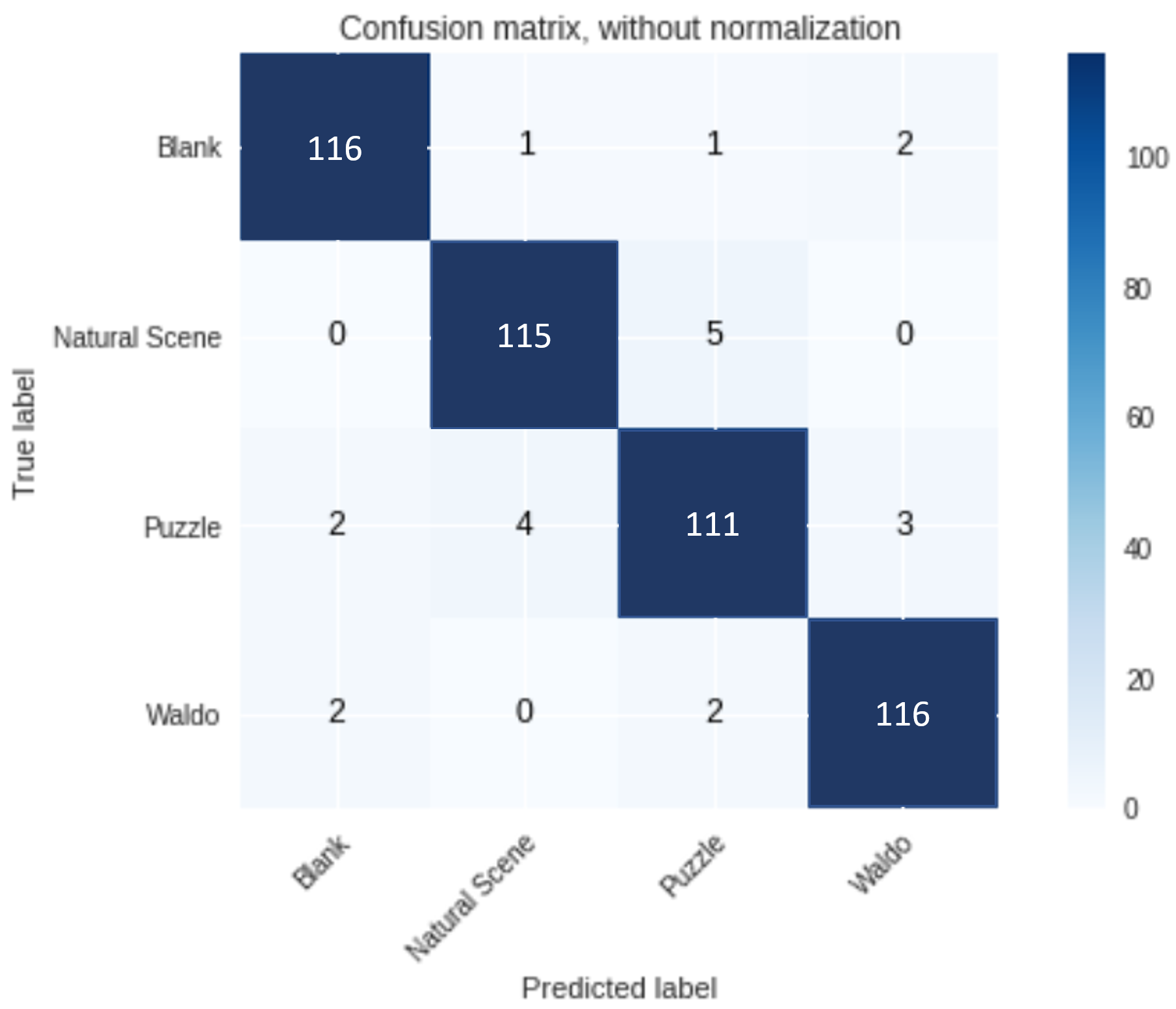}
  \caption{Confusion matrix of classification results from Ada Boosting with Decision Trees.}
  \label{fig:conf_matrix}
\end{figure}

As shown in Table~\ref{tab:accuracy}, Ada Boosting with a base classifier as decision tree performed better than SVM in this case. The decision tree separates the data based on attributes to multiple dimensions whereas SVM uses a nonlinear kernel to map the data to higher dimensions. Since a decision tree classifier with boosting works better for this dataset, it suggests that the variables \textit{LX Pix, LY Pix, LX href, LY href, and LP} can be separated based on some intervals of their values for each of the tasks. The detailed classification results can be seen in the form of a confusion matrix in Figure ~\ref{fig:conf_matrix}. The least number of misclassified samples were from the \textit{Blank} task with just 4 users being misclassified. This is reasonable because every blank scene would have similar patterns for every user and it is easy to classify. The tasks \textit{Finding Waldo} and \textit{Natural Scene} both had 5 of the users misclassified followed by the task \textit{Puzzle} with most number of misclassifications counting to 8. This shows that the task \textit{Puzzle} is more specific to the task being performed as compared to the users' behavior of looking at a scene. Different puzzles require different ways to compare for each user and thus makes it harder for the classifier to predict the task based on just the user behavior.

\section{Conclusion}

In this paper, we have supported the hypothesis that it is possible to decode the task, which concludes that eye gazes from eye movement data, carry cognitive information such as a mental state which is highly related with the task the observer is carrying out. Using different classification algorithms we successfully predicted the observer's task from eye movement data. We used SVM and Ada Boosting classifiers with an accuracy of 80.3\% and 95.4\%, respectively. While our results are performing better than most of the classifiers used for the similar task, there is a possibility of obtaining better accuracy by selecting more relevant features and other classification algorithms in the literature. Also for this study, we removed the blinking samples from the dataset but it can be interesting to study the blinking patterns for future work. We will be running our classification model on a dataset from Green et al.~\cite{greene2012reconsidering} and Yarbus~\cite{yarbus2013eye} to support their claim too and compare the accuracy.

\section*{Acknowledgments}
This research was partially supported by NSF grant IIS 1527200 and MSIT, Korea, under the ICT Consilience Creative program (IITP-2019-H8601-15-1011) supervised by the IITP.
%
\bibliographystyle{ACM-Reference-Format}
\bibliography{sample-base}

%


\end{document}